%% file: Main.tex
\pdfoutput=1
\documentclass[11pt]{article}
\usepackage[final]{acl}

\usepackage{times}
\usepackage{latexsym}
\usepackage[T1]{fontenc}
\usepackage[utf8]{inputenc}
\usepackage{microtype}
\usepackage{inconsolata}
\usepackage{comment}
\usepackage{booktabs}  
\usepackage{colortbl} 
\usepackage{multirow} 
\usepackage{amssymb}
\usepackage{amsmath}
\usepackage{pdfpages}  
\usepackage{subcaption} 
\usepackage{graphicx}  
\usepackage{float}
\usepackage{hyperref}
\usepackage[whole]{bxcjkjatype}
\usepackage{amsmath}
\usepackage{booktabs}

\usepackage{subcaption} 
\usepackage{listings}

\title{How Well Can LLMs Simulate Real Learner Evaluations \\of Educational Feedback?}

\author{
Momoka Furuhashi${}^{1,2}$
Kouta Nakayama${}^{2}$ 
Takashi Kodama${}^{2}$ 
Saku Sugawara${}^{3,4}$ 
Kyosuke Takami${}^{5}$ \\
${}^{1}$Tohoku University \hspace{1em}
${}^{2}$Research and Development Center for Large Language Models,\\
National Institute of Informatics \hspace{1em}
${}^{3}$National Institute of Informatics \\
${}^{4}$University of Tokyo \hspace{1em}
${}^{5}$Osaka Kyoiku University\\
\texttt{furuhashi.momoka.p4@dc.tohoku.ac.jp} \hspace{1em}
\texttt{\{nakayama,tkodama,saku\}@nii.ac.jp}\\
\texttt{takami-k75@cc.osaka-kyoiku.ac.jp}
}

\begin{document}
\maketitle
\begin{abstract}
While recent studies have explored human behavior and preference simulation using large language models (LLMs), it remains unclear how well LLMs can simulate subjective evaluations from real learners in educational settings.
We investigate this question using real learner evaluation data on feedback for high-school biology questions at both the group and individual levels. 
We compare performance with and without learner-specific information, such as personality traits and evaluation examples, across six models.
Our results show that LLMs still have a limited ability to simulate learner evaluations.
Providing learner profiles and examples improves score calibration and individual-level simulation, but more often fails to improve group-level consistency.
These findings highlight the need to investigate which learner information and adaptation strategies are effective for learner preference simulation.

\end{abstract}

\section{Introduction}
\label{sec:introduction}

Recent advances in large language models (LLMs) have inspired growing interest in simulating human behavior and preferences. 
Prior studies investigate persona-based role-playing and human-like behavior generation~\cite{hu-collier-2024-quantifying,samuel-etal-2025-personagym}, while more recent work examines group-level human simulation~\cite{hu2026simbench}.

This direction is particularly important in education, where empirical studies with real learners are costly and time-consuming.
In addition, learner populations vary widely in factors, such as prior knowledge and learning attitudes, which further complicates large-scale educational studies with real learners.
To address these challenges, studies explore LLM-based learner simulation as a substitute for diverse learner populations~\cite{liu-etal-2024-personality,sanyal-etal-2025-investigating,martynova-etal-2025-llms}. However, these studies do not examine whether LLM behaviors align with those of real learners.

One important application of learner preference simulation is educational feedback evaluation, where effectiveness depends on learners' subjective preferences.
Feedback tailored to learner responses improves learning outcomes~\cite{Hattie2007,wisniewski2020power}, and recent studies use LLMs to generate and evaluate feedback~\cite{tonga2025simulatingllmtollmtutoringmultilingual,qian2025deanllmtutorsexploring,chu2026feedevalpedagogicallyalignedevaluation}. 
However, evaluating feedback quality requires comparisons with real learner evaluations (RLEs), and it remains unclear how well LLMs can simulate such subjective evaluations.

\begin{figure*}[t]
\includegraphics[width=\textwidth]{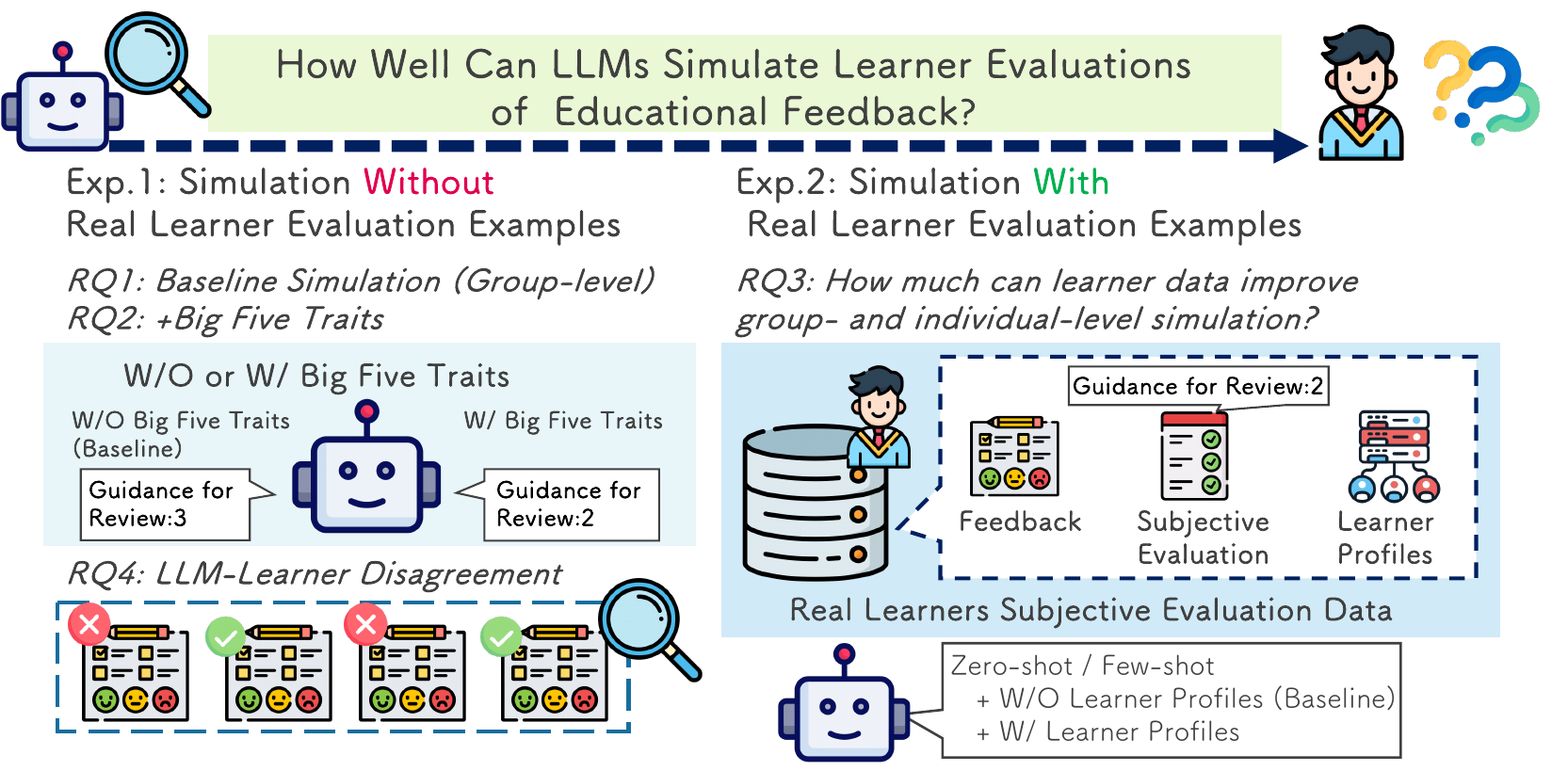}
\caption{Overview of the experiments and research questions. 
We first evaluate LLM simulation without learner-specific evaluation data (Exp.~1), and then examine performance using real learner evaluation data, such as profiles and evaluation examples (Exp.~2). 
At the group level, we investigate simulation performance (RQ1) the effects of Big Five traits (RQ2), and the contribution of learner evaluation data to group- and individual-level simulation (RQ3). Finally, we analyze feedback types that lead to disagreement between LLMs and learner evaluations (RQ4).}
\label{fig:overview}
\end{figure*}

We investigate how well LLMs can simulate subjective learner evaluations using a RLEs dataset from~\citet{furuhashi2026investigatinglearnerawaredesignllmgenerated}, which contains three-point evaluations on six criteria for feedback on learners' answers to high-school biology questions. 
In practice, learner preference simulation is often required without learner-specific information, while richer learner data may enable more personalized simulation.
Figure~\ref{fig:overview} shows the experimental design and research questions (RQs).
RQ1:~\textit{How well can LLMs simulate group-level subjective evaluation tendencies without profiles?}
RQ2:~\textit{How much can Big Five personality traits improve group-level learner preference simulation?}
RQ3:~\textit{How well can LLMs simulate group- and individual-level learner evaluation tendencies when real learner information is available?}
RQ4:~\textit{What feedback characteristics lead to LLM-learner disagreement?}

To investigate RQ1, RQ2, and RQ3, we conduct group-level simulation experiments using six models under two settings: (i) a setting without RLEs (§~\ref{ssec:experiment_1}) and (ii) a setting using RLEs data, including learner profiles and evaluation examples (§~\ref{ssec:experiment_2}). 
We then compare how accurately each setting simulates learner-group evaluation tendencies. 
We also analyze how well LLMs can simulate learner-specific evaluation tendencies using RLEs for RQ3 (§~\ref{ssec:individual_results}).
To investigate RQ4, we analyze feedback instances that show disagreements between LLM and learner evaluations and examine their characteristics (§~\ref{ssec:fb_type_collect_results}).
We evaluate simulation performance using both Spearman's $\rho$, which measures ranking consistency, and MAE, which measures score-level agreement.

Our experiments yield four key findings. 
First, LLMs still show limited reliability in simulation learner evaluations at both the group and individual levels. 
Second, at the group level, large-scale models improve score-level agreement when learner profiles and evaluation examples are provided, but their ability to preserve evaluation consistency across feedback instances often decreases.
In contrast, smaller models tend to benefit more from learner profiles in terms of evaluation consistency.
Third, at the individual level, learner profile and evaluation examples generally improve evaluation agreement across all models. 
Fourth, LLMs tend to disagree with learners on highly detailed and symbol-heavy feedback. 
These findings highlight the need to investigate how learner information, prompting strategies, and model adaptation methods can improve learner preference simulation.

Our contributions are summarized as follows:
\begin{itemize}
    \item We investigate how well LLMs simulate learner evaluations of educational feedback at group and individual levels using RLEs.
    \item We show that learner profiles improve score-level agreement and individual-level simulation, but often do not improve evaluation consistency at the group level.
    \item We analyze disagreement patterns between LLM and learner evaluations, including LLM overestimation of highly detailed and symbol-heavy feedback, and discuss implications for learner information and adaptation strategies.
\end{itemize}

\section{Related Work}
\label{sec:related_works}

\subsection{LLM-based Human Simulation}
\label{ssec:llm-based_simulation}
Interest in  whether LLMs can imitate human behavior and preference has recently grown. 
Early studies examine persona-based role-playing using personality traits or profile attributes as prompts~\cite{huang-hadfi-2024-personality,jiang-etal-2024-personallm,samuel-etal-2025-personagym,xie2025human}.
\citet{chen2026systematicanalysisimpactpersona} further show that personality conditioning can induce behavioral tendencies that are consistent with human personality–cognition relationships, although the effects vary across tasks.
Recent work extends this direction from persona imitation to human behavior simulation and personalized preference modeling.
\citet{hu2026simbench} evaluate whether 45 models can simulate human groups across diverse datasets and report that even strong models do not provide consistently reliable human simulation. \citet{ma2026personalizedrewardbenchevaluatingreward} evaluate whether reward models can capture individual-specific preferences and show that state-of-the-art models still struggle with personalization.
Building on these lines of work, we investigate how well LLMs can simulate subjective evaluation patterns in the educational domain at both the group and individual levels.

\subsection{LLM-Based Learner Simulation}
\label{ssec:education}

Research on learner simulation with LLMs is also advancing in education~\cite{chu-etal-2025-llm}. Prior studies use simulated learners to evaluate the quality of LLM-generated hints and feedback~\cite{tonga2024automaticgenerationquestionhints,tonga2025simulatingllmtollmtutoringmultilingual}, generate learner dialogues~\cite{martynova-etal-2025-llms}, and simulate diverse learning styles and personality traits~\cite{liu-etal-2024-personality,sanyal-etal-2025-investigating}. However, these studies generally do not examine whether simulated behaviors align with evaluations from real learners.
Several studies also use real learner data. \citet{zhu2025edupersonabenchmarkingsubjectiveability} simulate learner dialogues based on personality traits, while \citet{asano-etal-2025-llms} investigate whether LLMs can imitate learners' solution patterns in open-ended mathematics problems. 
However, both studies focus on dialogue generation or solution generation rather than subjective feedback evaluations.
We focus on simulating subjective feedback evaluations using real learner data.

\subsection{Feedback Generation and Evaluation}
\label{ssec:feedback}
Feedback effectively supports learning by helping learners bridge gaps between their answers and correct answers~\cite{Hattie2007,wisniewski2020power}. 
Many studies have used LLMs to automatically generate feedback for learner responses, and such approaches have attracted attention as a way to replace or support teachers~\cite{nair-etal-2024-closing,zhao-etal-2025-learnlens,chu-etal-2025-llm}.
Recent work also investigates pedagogically grounded feedback generation and evaluation.
\citet{borges-etal-2024-teach} propose a taxonomy of educational feedback based on pedagogical elements, and \citet{furuhashi2026investigatinglearnerawaredesignllmgenerated} use part of this taxonomy to generate six types of feedback for high-school biology questions.
Other studies use LLMs as evaluators to assess the quality of generated feedback.
\citet{qian2025deanllmtutorsexploring} evaluate feedback for computer science assignments on content quality,  effectiveness, and hallucination, while \citet{chu2026feedevalpedagogicallyalignedevaluation} evaluate essay feedback on specificity and helpfulness. 
However, these studies rely on LLM-based or predefined evaluations and do not examine alignment with subjective evaluations from real learners. 
Since feedback evaluations depend on learner profiles, we investigate how well LLMs can simulate learner evaluations.

\section{Dataset}
\label{sec:dataset}

\begin{table}[t]
\centering
\small
\input{table/feedback_type}
\caption{Overview of the six feedback prompts from the dataset of \citet{furuhashi2026investigatinglearnerawaredesignllmgenerated}, with each prompt's name and explanation.}
\label{tb:feedback_type}

\end{table}

We use the dataset from \citet{furuhashi2026investigatinglearnerawaredesignllmgenerated}, which contains students' responses, their evaluations of LLM-generated feedback for 
high-school biology multiple-choice questions, and their profiles information collected through a pre-questionnaire.
The dataset includes seven questions (37 options in total), some of which involve images.
For each option, they generate feedback using LLMs with six prompts listed in Table~\ref{tb:feedback_type}: Normal, Keywords, Actionability, Novelty, Coverage, and Positivity, resulting in a total of 222 feedback instances.
A total of 321 first-year high school students in Japan answered these questions and evaluated the feedback corresponding to their answers.
They rated each instance on a three-point scale across six criteria, such as clarity and ease of understanding, regardless of whether their answers were correct or incorrect.
Since students iteratively answer questions and evaluate feedback until they reach the correct answer, the dataset contains multiple evaluations for the same questions.
In total, the dataset includes 4,478 feedback evaluations.\footnote{Each feedback instance is evaluated across six criteria, resulting in a total of 26,868 evaluation scores.}
They randomly assigned the type of feedback to each student-question pair.
They also completed a pre-questionnaire concerning their background and prior experience.
The questionnaire included items about their interest in and confidence regarding biology learning and previous use of generative AI tools for learning or other purposes.
Example items include  ``I am good at basic biology. (1: Very poor;  5: Very good)'' and ``I think AI-generated feedback is useful. (1: Not useful at all; 5: Very useful)''.
The questionnaire also assessed personality trait information using a Japanese 70-item binary-choice (``yes'' / ``no'') Big Five personality inventory~\citep{Murakami1997BigFive} measuring five dimensions: Openness, Conscientiousness, Extraversion, Agreeableness, and Neuroticism.
The inventory includes items reflecting tendencies such as being curious about new ideas, being organized, and enjoying social interaction.
See Appendix~\ref{app:fb_info}.

\section{Experiments}
\label{sec:experiments}
We conduct two experiments to examine whether LLMs can simulate learners' subjective evaluations at the group level. 
First, we investigate evaluation simulation without RLEs data (§~\ref{ssec:experiment_1}). 
We then examine to how much learner profiles and evaluation examples improve alignment with learner evaluations when RLEs are available (§~\ref{ssec:experiment_2}). 
These experiments reveal the potential and limitations of evaluation simulation with LLMs.

\subsection{Experimental Setup}
We assess group-level evaluation simulation by aggregating LLM-generated evaluation scores for each feedback type and comparing them with RLEs.
We use six models: gpt-5-2025-08-07 (GPT-5)~\cite{GPT-5}, gemini-3.5-flash (Gemini-3.5-flash)~\cite{gemini3.5flash2026}, gemini-2.5-flash (Gemini-2.5-flash)~\cite{comanici2025gemini25pushingfrontier}, gemma-4-31B-it (Gemma-4-31B-it)~\cite{gemmateam2026gemma4}, Qwen3-VL-32B-Instruct (Qwen3-VL-32B-it), and Qwen3-VL-8B-Instruct (Qwen3-VL-8B-it)~\cite{bai2025qwen3vltechnicalreport}.
We consider two conditions based on whether RLEs are provided. 
To evaluate alignment between LLMs and human evaluations, we use two metrics: \textbf{Spearman's rank correlation coefficient} for ranking consistency and \textbf{Mean Absolute Error (MAE)} for score differences.
Scores are aggregated by feedback type, and the metrics are computed over mean scores across 36 combinations of evaluation criteria and feedback types.

\begin{table}[t]
\centering
\small
\resizebox{\columnwidth}{!}{
\input{table/pre_results_overall}
}
\caption{Agreement between LLMs and learner group evaluations measured by Spearman's $\rho$ and MAE.
Baseline achieves higher Spearman correlation in some cases, whereas adding Big Five information improves MAE.}
\label{tb:pre_results_overall}
\end{table}

\subsection{Experiments 1: Learner Preference Simulation Without RLEs Examples}
\label{ssec:experiment_1}
We examine whether LLMs can simulate the evaluation tendencies of a learner group without RLEs.

\paragraph{Without Big Five Personality Traits (Baseline)}
To investigate~\textit{RQ1: How well can LLMs simulate group-level subjective evaluation tendencies without profiles}, we use a baseline setting in which LLMs evaluate all 222 feedback instances without persona information.
We conduct three evaluation trials for each feedback instance and use the average score as the prediction. 
We then aggregate predictions by feedback type and evaluation criterion to obtain group-level tendencies. 
It allows us to examine how closely the default tendencies of LLMs align with those of a specific learner group, without explicitly imitating individual learners.

\begin{figure*}[t]
    \centering

    \begin{subfigure}[t]{\textwidth}
        \centering
        \includegraphics[width=\textwidth]{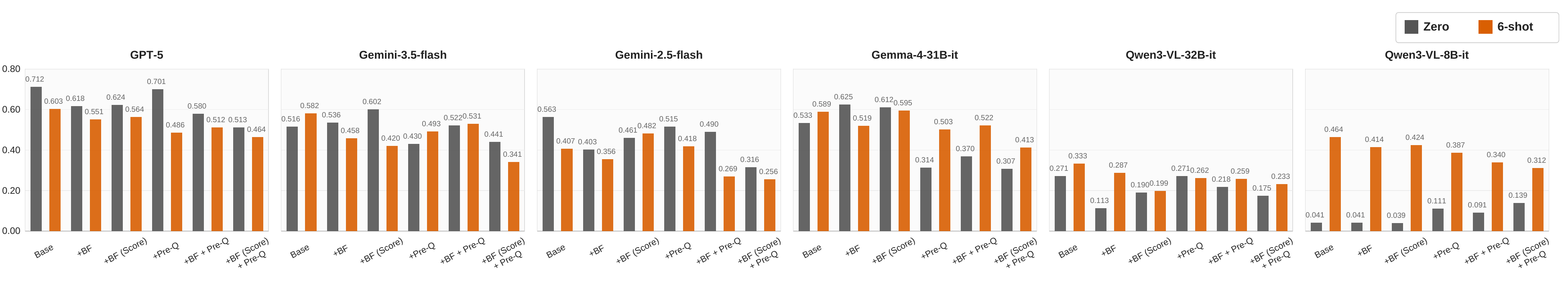}
        \caption{Spearman's $\rho$}
        \label{fig:spearman_barplots}
    \end{subfigure}


    \begin{subfigure}[t]{\textwidth}
        \centering
        \includegraphics[width=\textwidth]{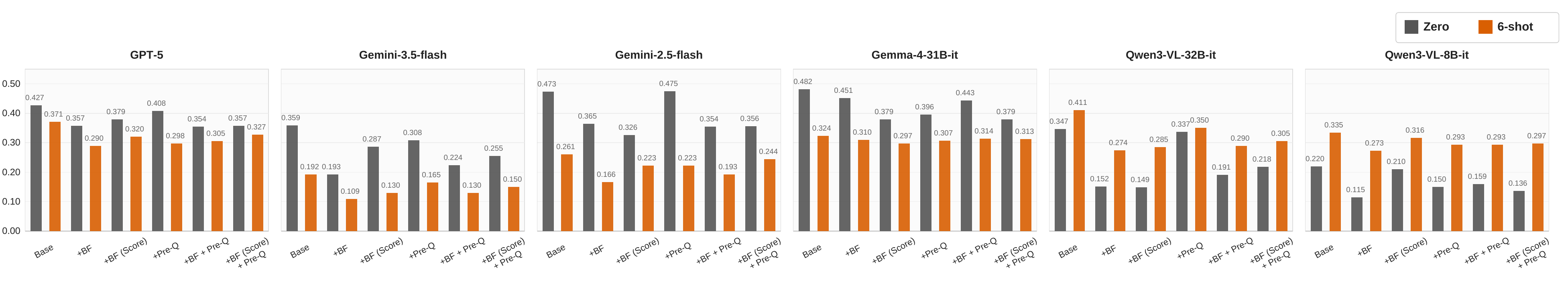}
        \caption{MAE}
        \label{fig:mae_barplots}
    \end{subfigure}
    \caption{Group-level simulation results across models for Spearman's $\rho$ (\ref{fig:spearman_barplots}) and MAE (\ref{fig:mae_barplots}). 
Larger models often achieve strong Spearman's $\rho$ under the zero-shot Baseline condition.
The effect of few-shot prompting on Spearman's $\rho$ varies by model scale.
For MAE, Big Five trait conditions, especially +BF and +BF (Score), generally reduce errors relative to the Baseline.
Few-shot prompting improves MAE for GPT-5, Gemini, and Gemma models, whereas it tends to increase errors for the Qwen3-VL models.}
    \label{fig:overall_results}
\end{figure*}

\paragraph{With Big Five Personality Traits（+BF）}
To investigate~\textit{RQ2: How much can Big Five personality traits improve group-level learner preference simulation?}, we introduce a setting that reflects the personality distribution of the learner population.
Because evaluating all combinations of feedback instances and trait combinations would be computationally expensive, we randomly sample 20 set of Big Five traits from the dataset for each feedback instance.
We append these traits to the prompt and generate persona-conditioned evaluations (4,440 samples in total).
We then aggregate the evaluations to examine whether traits improves group-level simulation.
We conduct one evaluation trial for each persona-conditioned prompt in this setting.

\paragraph{Result 1: Big Five Traits Do Not Consistently Improve Alignment}
\label{ssec:results_1}

We find that personality traits improve MAE but do not necessarily improve Spearman's $\rho$ across all models (see Table~\ref{tb:pre_results_overall}). 
Higher Spearman's $\rho$ indicates better preservation of relative evaluation tendencies across feedback instances, whereas lower MAE indicates closer agreement with human evaluations.
Personality information tends to improve score-level agreement by shifting evaluation scores closer to the human evaluation range, but does not consistently preserve relative evaluation tendencies between feedback instances.
In terms of Spearman's $\rho$, while personality information is effective for Gemini-3.5-flash ($0.434 \rightarrow 0.706$), Gemini-2.5-flash shows a substantial decrease ($0.532 \rightarrow 0.098$), suggesting that adding personality information can destabilize evaluation behavior for certain models.
We also find that large-scale models can simulate evaluation tendencies even without personality information, whereas smaller models show limited simulation performance regardless of personality traits.
Overall, these results suggest that preserving relative evaluation tendencies depends on the specific model, while personality information is more effective for improving score-level agreement.

\begin{figure*}[t]
    \centering

    \begin{subfigure}[t]{\textwidth}
        \centering
        \includegraphics[width=\textwidth]{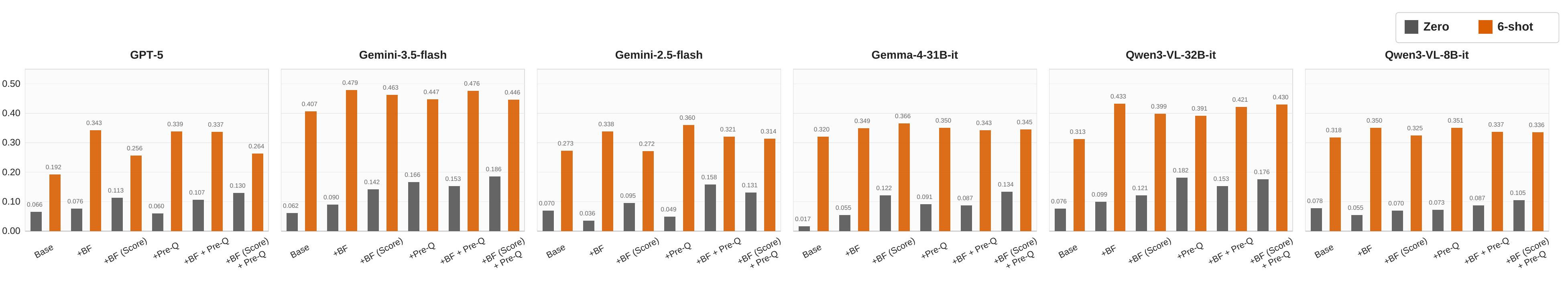}
        \caption{Spearman's $\rho$}
        \label{fig:spearman_barplots_individual}
    \end{subfigure}


    \begin{subfigure}[t]{\textwidth}
        \centering
        \includegraphics[width=\textwidth]{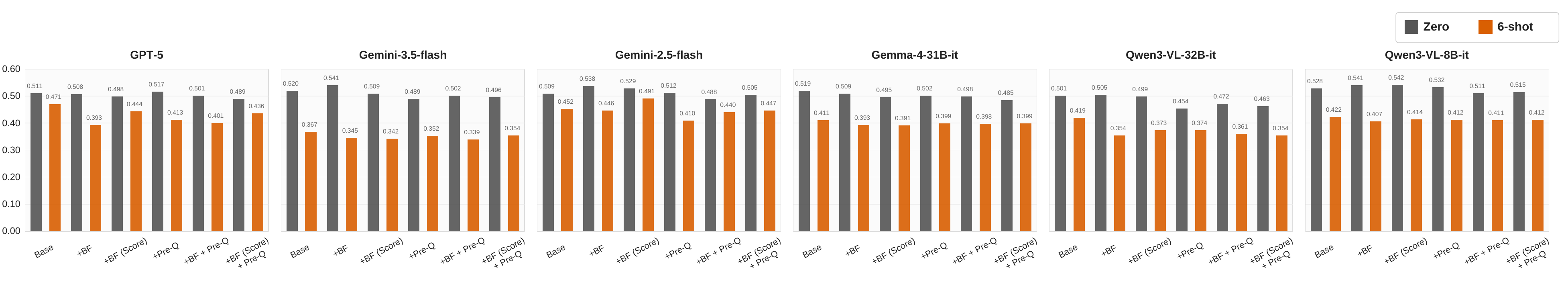}
        \caption{MAE}
        \label{fig:mae_barplots_individual}
    \end{subfigure}
    \caption{Individual-level simulation results across models for Spearman's $\rho$ (\ref{fig:spearman_barplots_individual}) and MAE (\ref{fig:mae_barplots_individual}). 
For Spearman's $\rho$, zero-shot settings often achieve correlations around 0.1, whereas Few-shot settings substantially improve performance across all cases. 
In many cases, conditions with learner profile information outperform the Baseline condition. 
For MAE, Few-shot settings consistently achieve lower errors than zero-shot settings across all models.}
    \label{fig:individual_results}
\end{figure*}

\subsection{Experiments 2: Learner Preference Simulation With RLEs Examples}
\label{ssec:experiment_2}

To investigate the group-level aspect of RQ3, we examine simulation performance when privileged RLEs, such as evaluation examples and profiles, is available.
Unlike §~\ref{ssec:experiment_1}, where answer choices are uniformly sampled, we use the actual distribution of learner responses to maintain comparability with RLEs.
We adopt two settings: a zero-shot setting and a few-shot setting that provides six evaluation examples from the same learner on different samples as context.
For each setting, we consider six conditions: \textbf{Baseline} (without learner information), \textbf{+BF} (Big Five questionnaire responses described in text), \textbf{+BF(Score)} (numerical Big Five personality scores), \textbf{+Pre-Q} (responses to pre-questionnaires), \textbf{+BF+Pre-Q}, and \textbf{+BF(Score)+Pre-Q}. 
These conditions allow us to analyze how different forms of learner information contribute to evaluation simulation.
We introduce +BF(Score) to examine whether differences between +BF and +BF(Score) arise from the information format or prompt length.
We use 658 evaluation samples randomly selected from the first evaluation completed by each learner.
The detailed prompts are provided in Figures~\ref{fig:basic} --~\ref{fig:pre_questionnaire_and_bigfive_score}.

\paragraph{Zero-shot and Few-shot (In-context Learning)}
\label{sssec:zeroshot}
We provide only learner profiles without evaluation examples. 
This setting allows us to evaluate the contribution of each type of contextual information independently.
We introduce a few-shot setting by providing evaluation examples from each learner through their evaluations on the other six questions.
This setting allows LLMs to infer learner-specific evaluation tendencies through in-context learning and examine how much such information improves the simulation of evaluation diversity.
The detailed prompts are provided in Figures~\ref{fig:feedback_evaluation} and~\ref{fig:feedback_evaluation_fewshot}.

\begin{figure*}[t]
    \centering

    \begin{subfigure}[t]{0.48\linewidth}
        \centering
        \includegraphics[width=\linewidth]
        {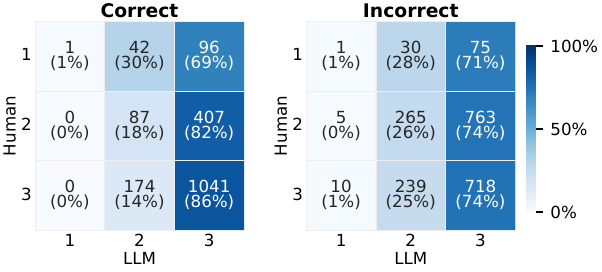}
        \caption{GPT-5}
        \label{fig:gpt5_cm}
    \end{subfigure}
    \hfill
    \begin{subfigure}[t]{0.48\linewidth}
        \centering
        \includegraphics[width=\linewidth]
        {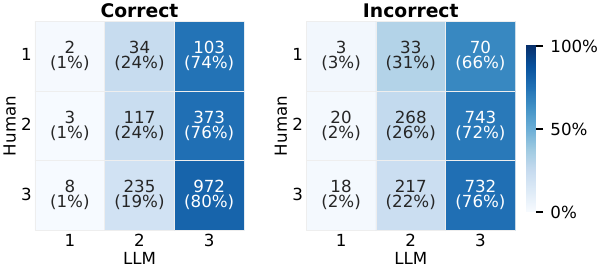}
        \caption{Gemini-3.5-flash}
        \label{fig:gemini3.5_cm}
    \end{subfigure}


    \begin{subfigure}[t]{0.48\linewidth}
        \centering
        \includegraphics[width=\linewidth]
        {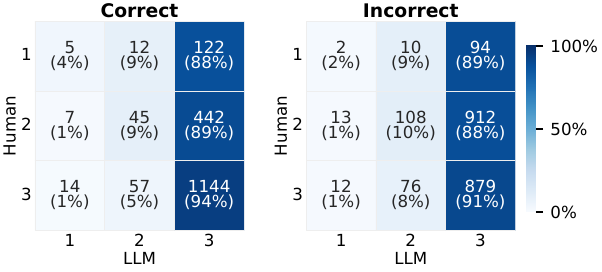}
        \caption{Gemini-2.5-flash}
        \label{fig:gemini2.5_cm}
    \end{subfigure}
    \hfill
    \begin{subfigure}[t]{0.48\linewidth}
        \centering
        \includegraphics[width=\linewidth]
        {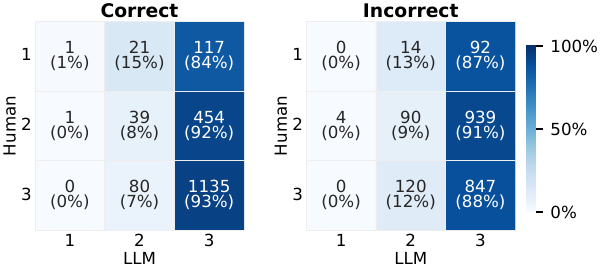}
        \caption{Gemma-4-31B-it}
        \label{fig:gemma_cm}
    \end{subfigure}

    \begin{subfigure}[t]{0.48\linewidth}
        \centering
        \includegraphics[width=\linewidth]
        {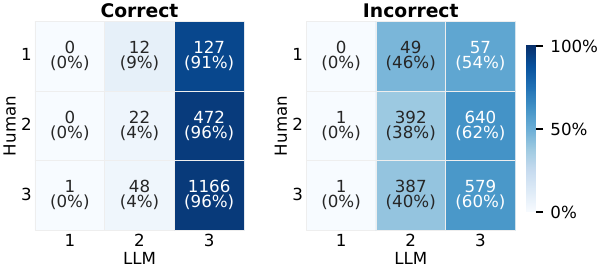}
        \caption{Qwen3-VL-32B-it}
        \label{fig:qwen32_cm}
    \end{subfigure}
    \hfill
    \begin{subfigure}[t]{0.48\linewidth}
        \centering
        \includegraphics[width=\linewidth]
        {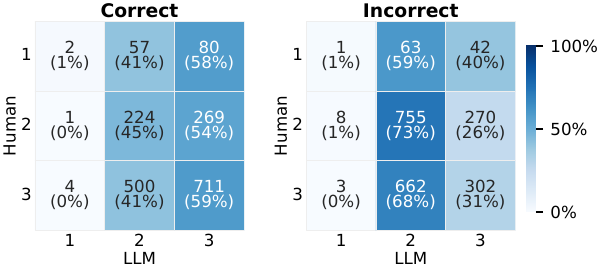}
        \caption{Qwen3-VL-8B-it}
        \label{fig:qwen8_cm}
    \end{subfigure}

    \caption{Confusion matrices between human and LLM evaluations (zero-shot Baseline setting). 
Each subplot compares rating distributions for correct and incorrect answers.
Large-scale models tend to assign overly positive ratings for correct answers, whereas smaller models show stricter or middle-range ratings for incorrect answers.}
    \label{fig:confusion_matrix}
\end{figure*}

\paragraph{Result 2: More Context Does Not Always Help}
\label{ssec:results_2}

We find that additional information and few-shot setting do not consistently improve learner preference simulation. 
For Spearman's $\rho$, large-scale models tend to achieve the highest performance under the Baseline condition in both zero-shot and few-shot settings (Figure~\ref{fig:spearman_barplots}). 
The effects of few-shot setting also differ by model size. 
For large-scale models, few-shot setting often fails to improve and sometimes decreases Spearman's $\rho$, whereas smaller models show substantial improvements. 
Adding learner information does not necessarily improve performance, even when multiple profiles are combined.
These results suggest that large-scale models may already encode an implicit representation of an ``average learner'' without additional information, and that evaluation examples may instead disturb evaluation tendencies.
In contrast, smaller models appear to benefit more from evaluation examples than from learner profiles.

For MAE, personality-based conditions (+BF and +BF(Score)) often improve performance (Figure~\ref{fig:mae_barplots}).
Large-scale models also tend to achieve lower MAE under few-shot settings, whereas smaller models do not show consistent improvements. 
Overall, model without profiles better preserve evaluation tendencies for several models, whereas this trend is not observed for the Qwen3-VL models.

\section{Analysis}
\label{sec:analysis}
We investigate individual-level simulation performance and group-level disagreement patterns across feedback types using the data from §\ref{ssec:experiment_2}.

\subsection{Individual Preference Simulation}
\label{ssec:individual_results}

To investigate the individual-level aspect of RQ3, we analyze how well LLMs simulate learner-specific evaluation tendencies.
We find that LLMs still struggle to capture such preferences, although few-shot settings consistently improve performance.
We measure agreement using Spearman's $\rho$ and MAE, where each (learner, question, feedback, criterion) tuple is treated as a single observation.
Unlike group-level evaluation, metrics are computed directly from paired ratings without aggregation.
Figures~\ref{fig:spearman_barplots_individual} and~\ref{fig:mae_barplots_individual} show the results for Spearman's $\rho$ and MAE, respectively.
Under the zero-shot setting, Spearman $\rho$ remain around $0.1$ for many models. 
This result indicates that LLMs struggle to simulate individual-level preferences. 
In contrast, the few-shot setting improves correlations across all cases.
In particular, Gemini-3.5-flash achieves a maximum correlation of $0.479$ (+BF). 
We also observe cases where Qwen3-VL-8B-it achieves correlations comparable to or higher than those of larger models.
For MAE, all models show lower errors under the few-shot setting, consistent with the results in §~\ref{ssec:results_2}. 
Learner profiles and evaluation examples improve score calibration even at the individual level.
Unlike the group-level results, settings with learner information often outperform the Baseline setting, although no learner profile consistently improves performance across models. 
However, correlations remain low in many cases, and LLMs still have limited ability to simulate individual-level evaluation tendencies.

\subsection{Correctness Affects Evaluation Simulation}
\label{ssec:correct_results}

Answer correctness may influence subjective evaluations of feedback at the group-level. 
Learners may evaluate feedback more positively for correct answers and more strictly for incorrect answers. 
To examine whether LLMs simulate these tendencies or simply rely on answer correctness, we focus on the zero-shot Baseline settings, as it provides the cleanest view of the model's inherent behavior without personalization.
We divide the results into correct-answer and incorrect-answer groups and compare agreement with human evaluations using confusion matrices, Spearman's $\rho$, and MAE 
(see Figure~\ref{fig:confusion_matrix} and
Table~\ref{tb:spearman}).
Figure~\ref{fig:confusion_matrix} reveals different tendencies across model scales.
\begin{table}[t]
\centering
\small
\resizebox{\columnwidth}{!}{
\input{table/Spearman}
}
\caption{Spearman's $\rho$ and MAE for overall, correct-answer, and incorrect-answer groups under the zero-shot Baseline setting. Large-scale models show higher agreement for incorrect answers, whereas smaller models tend to show higher agreement for correct answers.}
\label{tb:spearman}
\end{table}
\begin{table*}[t]
\centering
\small
\resizebox{\textwidth}{!}{
\input{table/analysis_fb_type_collect}

}
\caption{Distribution of large disagreement cases ($|$difference$| \geq 1.0$) across feedback types.
Values indicate cases counts for each model. 
Because each feedback instance is evaluated on six criteria, disagreements are counted separately by criterion.
High and Low indicate cases where the LLM assigns higher or lower ratings than humans.}
\label{tb:analysis_fb_type_collect}
\end{table*}
Large-scale models tend to assign overly positive evaluations in the correct-answer group, which limits agreement with human evaluations.
In the incorrect-answer group, their evaluations become stricter and reduce high scores.
In contrast, small-scale models tend to concentrate on middle-range scores even for correct-answers and fail to simulate the human evaluation distribution.
Their evaluations become even stricter in the incorrect-answer, which further increases disagreement with human evaluations.
Table~\ref{tb:spearman} also shows scale-dependent differences.
Large-scale models achieve the highest Spearman's $\rho$ on the overall data, followed by the incorrect-answer group, and also achieve the lowest MAE for incorrect answers.
These results suggest that large-scale models partially share correctness-dependent evaluation tendencies with learners.
In contrast, medium- and small-scale models achieve higher correlations for correct answers, while performance substantially decreases for incorrect answers.
This suggests that smaller models may over-adapt to incorrect answers because they assign uniformly strict evaluations.
As a result, agreement with human evaluation tendencies decreases. Even when MAE remains relatively low, Spearman's $\rho$ often deteriorates. 
This indicates that these models may match overall score distributions without reproducing fine-grained evaluation tendencies.

\subsection{Feedback Types Behind Disagreement}
\label{ssec:fb_type_collect_results}
\subsubsection{Disagreement Across Feedback Types}
Since the dataset contains six types of feedback, agreement between LLM and learner evaluations may vary depending on feedback types.
To investigate RQ4:~\textit{What feedback characteristics lead to LLM-learner disagreement?}, we focus on the zero-shot Baseline setting to analyze disagreement arising from the model's intrinsic evaluation behavior without personalization.
We extract cases where the absolute difference between the learner evaluation and the LLM evaluation exceeds 1.0 point.
Because each feedback instance is evaluated on six criteria, we count criteria separately. 
We classify cases where LLMs assign higher scores than learners as \textbf{High} and lower scores as \textbf{Low}.
Table~\ref{tb:analysis_fb_type_collect} shows the distribution of these disagreements across feedback types.
Across all models, LLMs tend to assign higher scores than learners to \textit{Coverage} instances with comprehensive information for reaching the correct answer.
Large- and medium-scale models tend to highly evaluate \textit{Actionability} instances with actionable instructions and \textit{Positivity} instances with positive expressions. 
In contrast, smaller model tends to highly evaluate \textit{Keywords} instances that emphasizes important terms.

\subsubsection{LLMs Overestimate Detailed Feedback}

\paragraph{Analysis Settings}
To identify what types of feedback lead to disagreements between LLM and human evaluations, we conduct a qualitative analysis on feedback instances that at 
least two models among all six models classify as High.
We analyze combinations of feedback instances and evaluation criteria that satisfy this condition (92 sets in total, corresponding to 33 feedback instances).

\paragraph{Comprehensive and Symbolic Feedback}
We find several common characteristics of feedback that LLMs tend to overestimate. 
We first observe LLMs favor feedback with high information coverage. 
This tendency frequently appears in \textit{Coverage}  (34 out of 92) and \textit{Actionability}  (16 out of 92) feedback. 
For feedback that describes detailed procedures or provides all information necessary to reach the correct answer, LLMs often assign higher scores than learners on multiple evaluation criteria, such as Key Points Clarity and Guidance for Review. 
These results suggest that comprehensive descriptions are strongly associated with higher text quality in LLM evaluation criteria, whereas large amounts of information do not necessarily improve usefulness for learners.
We also find LLMs highly evaluate feedback containing non-natural-language expressions, such as arrows ($\rightarrow$) and symbolic representations of concept relationships. 

\paragraph{Hallucinated Evaluation Reasons}
We also observe multiple cases where LLMs generate evaluation reasons based on hallucinated evidence that was not provided in the evaluation setting. 
Although the experiment provides no external information such as textbooks or teacher explanations, LLMs sometimes justify their evaluations using fabricated reasons such as ``consistent with the teacher's explanation'' or ``based on textbook knowledge.'' 
These results suggest that LLMs may rely on external educational priors rather than only the provided feedback content when generating evaluations.
See Appendix~\ref{app:disagreement_feedback_example} for details.

\subsection{Discussion}
Our results reveal limitations of LLM-based learner preference simulation at both the group and individual levels, consistent with prior findings on group-level human simulation~\cite{hu2026simbench}.
Similar challenges in capturing individual-specific preferences have also been reported for personalized reward models~\cite{ma2026personalizedrewardbenchevaluatingreward}.
Future work should explore several directions for improving learner preference simulation. 
First, prompt designs tailored for simulation tasks may improve LLMs' ability to simulate learner evaluations.
Beyond prompt engineering, it is also important to examine whether model adaptation methods, such as supervised fine-tuning (SFT) and reinforcement learning from human feedback (RLHF) can further improve simulation performance.
Second, richer learner information, such as pre/post-test performance, confidence in answers, and evaluation time, may help better capture learner-specific evaluation tendencies. Finally, our findings suggest the potential benefit of models designed specifically for educational simulation and learner modeling, beyond general-purpose instruction-following LLMs.

\section{Conclusion}
\label{sec:conclusion}
We investigate how well LLMs can simulate subjective evaluations by learners at both the group and individual levels using learner evaluation data on LLM-generated feedback for high-school biology questions. 
Our results show that LLMs still have a limited preference simulation ability at both levels. 
However, we also find that learner-specific information improves simulation performance at the individual level. 
This result highlights the need to better understand which learner information, prompting strategies, and model adaptation method are more effective for simulating learner-specific evaluation tendencies.

\section*{Limitations}
This study has several limitations.
First, this study has limited generalizability. 
We conduct preference simulation experiments only on feedback generated for high school biology tasks.
Therefore, it remains unclear whether our findings also apply to other subjects or educational domains. 
Future work should examine multiple subjects and diverse educational settings.
Second, this study uses limited learner information.
We use learners' demographic attributes and past evaluation histories, but many factors influence subjective evaluations in real educational settings, such as academic ability, motivation, interests, and prior learning experiences.
Therefore, the information used in this study may not fully capture learners' evaluation tendencies. Future work should incorporate a broader range of learner characteristics.
Third, subjective evaluations themselves may lack consistency. This study treats learners' subjective evaluations as ground-truth labels, but learners do not always share the same evaluation criteria.
Therefore, some discrepancies between LLM evaluations and learner evaluations may arise not only from limitations in LLM simulation performance but also from variability in human evaluations.
Finally, this study focuses only on subjective evaluations at a single time point.
We do not model dynamic learning processes in which learners change their understanding or evaluation criteria through feedback.
Future work should investigate learner preference simulation methods that incorporate long-term learning histories and learning processes.

\section*{Ethical considerations}
This study uses the learner evaluation dataset introduced in~\citet{furuhashi2026investigatinglearnerawaredesignllmgenerated}. The original data collection was conducted with institutional ethical approval and informed consent from participants. All data used in this study were anonymized, and no personally identifiable information was included in the experiments. 
Therefore, this study does not involve identifiable personal information or direct interaction with participants.
Further details are provided in Appendix~\ref{app:fb_info}.

\paragraph{Use of AI Assistants}
We used generative AI tools to assist with code debugging and modification, literature survey support, and English writing revision. 
The AI tools were used only as supportive assistants for improving readability and development efficiency, and all experimental designs, analyses, interpretations, and final manuscript contents were verified and finalized by the authors.

\section*{Acknowledgments}
The authors would like to thank the anonymous
reviewers for their helpful comments. 
This work was supported by 
JST FOREST Grant Number JPMJFR232R, 
JST BOOST Grant Number JPMJBS2421, JPMJBY24D9,
JSPS KAKENHI Grant Numbers JP23K17012, JP23K25698, and
partially funded by the Cabinet Office, Government of Japan, under the BRIDGE initiative through the MEXT measure ``Building Educational AI Models and Developing a Human-Agent Collaborative AI Ecosystem for Educational Support.''
In this work, we used the ``mdx: a platform for building data-empowered society.''
We thank Ryoko Tokuhisa for her constructive comments and suggestions that helped improve this paper.

\bibliography{references}
\appendix
\section{Dataset}
\label{app:fb_info}
In this section, we provide additional details about the dataset introduced by~\citet{furuhashi2026investigatinglearnerawaredesignllmgenerated} and used in §~\ref{sec:dataset}.
The dataset consists of feedback generated for biology questions answered by 321 first-year high school students in Japan. Each feedback instance was evaluated on a three-point scale across six criteria (See Appendix~\ref{app:criteria}).

\input{table/disagree_feedback_count}

\subsection{Participants}
The participating school is designated by the Japanese government as a Super Science High School (SSH)—a national program that supports selected schools in providing enhanced STEM education, including student-led research projects and formal instruction in research ethics. 
Students at SSH schools regularly conduct independent research and are therefore familiar with research practices, which likely fostered a sincere and conscientious attitude toward participation. 
They were approximately 15--16 years old and enrolled in academically selective high schools.

\subsection{Data Governance}
We obtained consent from the participants or guardians to use the anonymized data for research purposes and to release it as a dataset. 
All learner data provided to the LLM were anonymized in advance, and no personally identifiable information, such as names or student IDs, was included.
The LLM inputs consisted of anonymized response data, correctness information, Big Five questionnaire responses or derived scores, pre-survey information, and the selected feedback and its evaluation.
All of these data were used within the scope of the obtained consent.

\subsection{Evaluation Criteria}
\label{app:criteria}
Learners evaluated each feedback instance based on six criteria, using questions directly adopted from \citet{furuhashi2026investigatinglearnerawaredesignllmgenerated}.
\begin{itemize}
\item \textbf{Guidance for Review}: This feedback helped me understand what I should review.
\item \textbf{Trustworthiness}: I felt that the content of
this feedback was trustworthy.
\item \textbf{New Knowledge}: This feedback provided me with
new knowledge or perspectives.
\item \textbf{Key Points Clarity}: This feedback clarified the key points
needed to understand this task.
\item  \textbf{Ease of Understanding}: The explanation was easy to understand.
\item \textbf{Expression Quality}: I think the way this feedback was presented was good. 
\end{itemize}

\section{Qualitative Analysis of Disagreement Feedback Instances}
\label{app:disagreement_feedback_example}
\subsection{Results of Disagreement Feedback}
We analyze feedback instances where at least two of the six models assign ratings more than 1.0 point higher than the group-level learner evaluations.
As a result, 92 disagreement cases (corresponding to 33 feedback instances) are identified for analysis.
In contrast, only one case is observed where the LLM assigns a lower rating than learners despite a difference greater than 1.0 points.
Table~\ref{tb:disagree_feedback_count} shows the distribution across feedback types. 
\textit{Coverage} accounts for more than 35\% of all cases (34 out of 92), suggesting that LLMs tend to highly evaluate feedback with comprehensive information. 
We also observe that disagreement with human evaluations occurs more frequently for the Guidance for Review and Ease of Understanding criteria. 
For these criteria, \textit{Coverage}, \textit{Actionability}, and \textit{Positivity} feedback are particularly common.

\input{table/feedback_example1}
\input{table/feedback_example2}

\subsection{Example of Disagreement Feedback}
Tables~\ref{tb:fb_example1} and~\ref{tb:fb_example2} present representative feedback instances that produced large disagreements between group-level learner and LLM evaluations, together with the corresponding criteria.
In Table~\ref{tb:fb_example1}, the feedback is characterized by comprehensively covering the steps leading to the correct answer, and none of the models agreed with the human evaluations, as reflected in the evaluation reasons provided for each model.
In Table~\ref{tb:fb_example2}, the feedback tends to contain arrow notation in its content, and hallucinations were observed in some model outputs — for instance, references to statements such as ``the explanation was consistent with the teacher'' despite no such information being present in the feedback.

\section{Prompt}
\begin{figure*}[t]
\includegraphics[width=\textwidth]{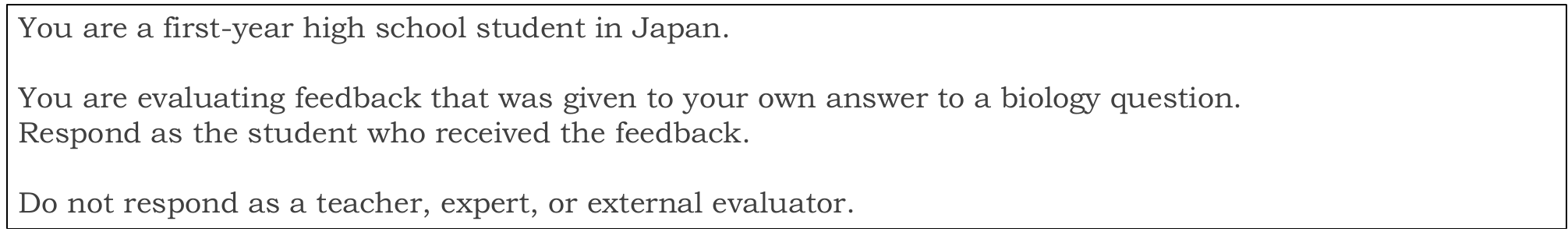}
\caption{Basic system prompt used as the baseline condition without learner profile information.
} 
\label{fig:basic}
\end{figure*}

\begin{figure*}[t]
\includegraphics[width=\textwidth]{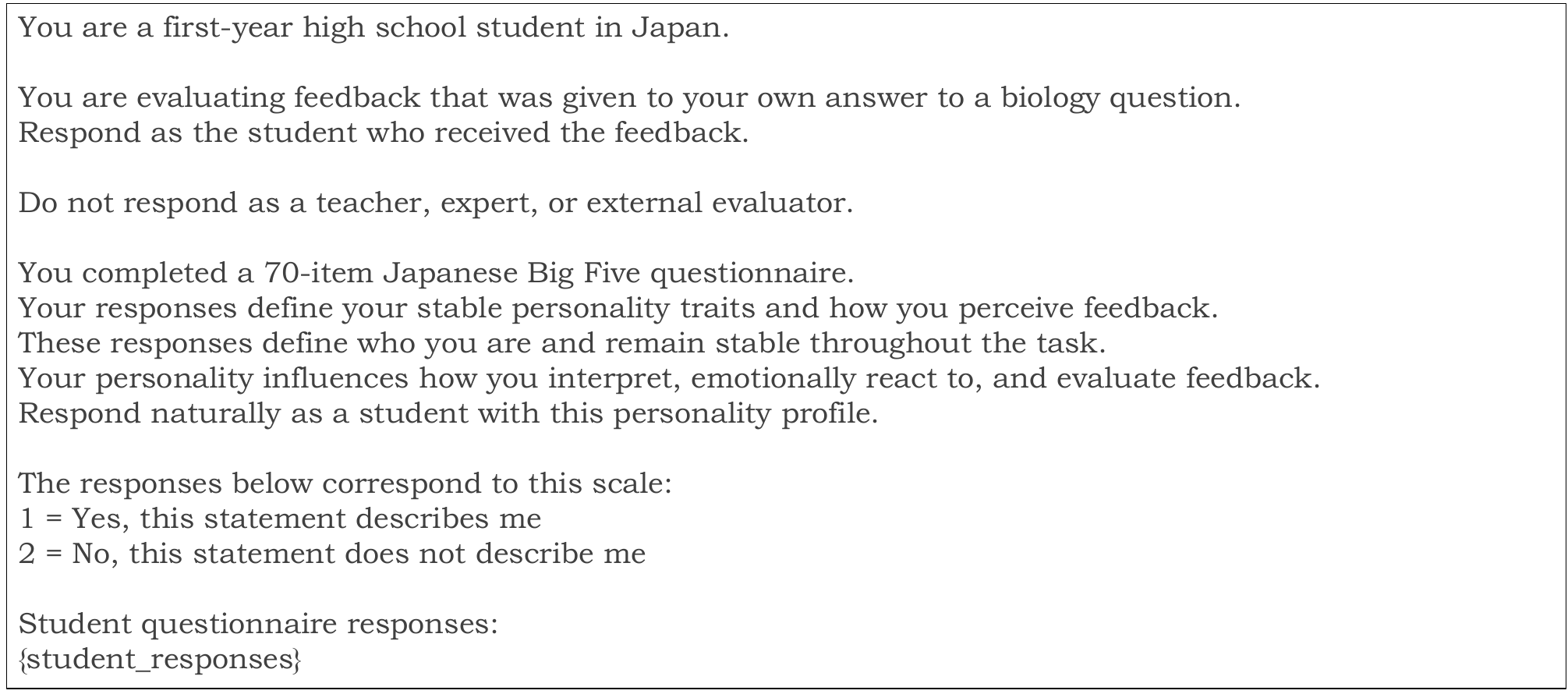}
\caption{System prompt using Big Five personality trait information based on the questionnaire proposed by~\citet{Murakami1997BigFive}, which consists of 70 items.
} 
\label{fig:all_bigfive}
\end{figure*}

\begin{figure*}[t]
\includegraphics[width=\textwidth]{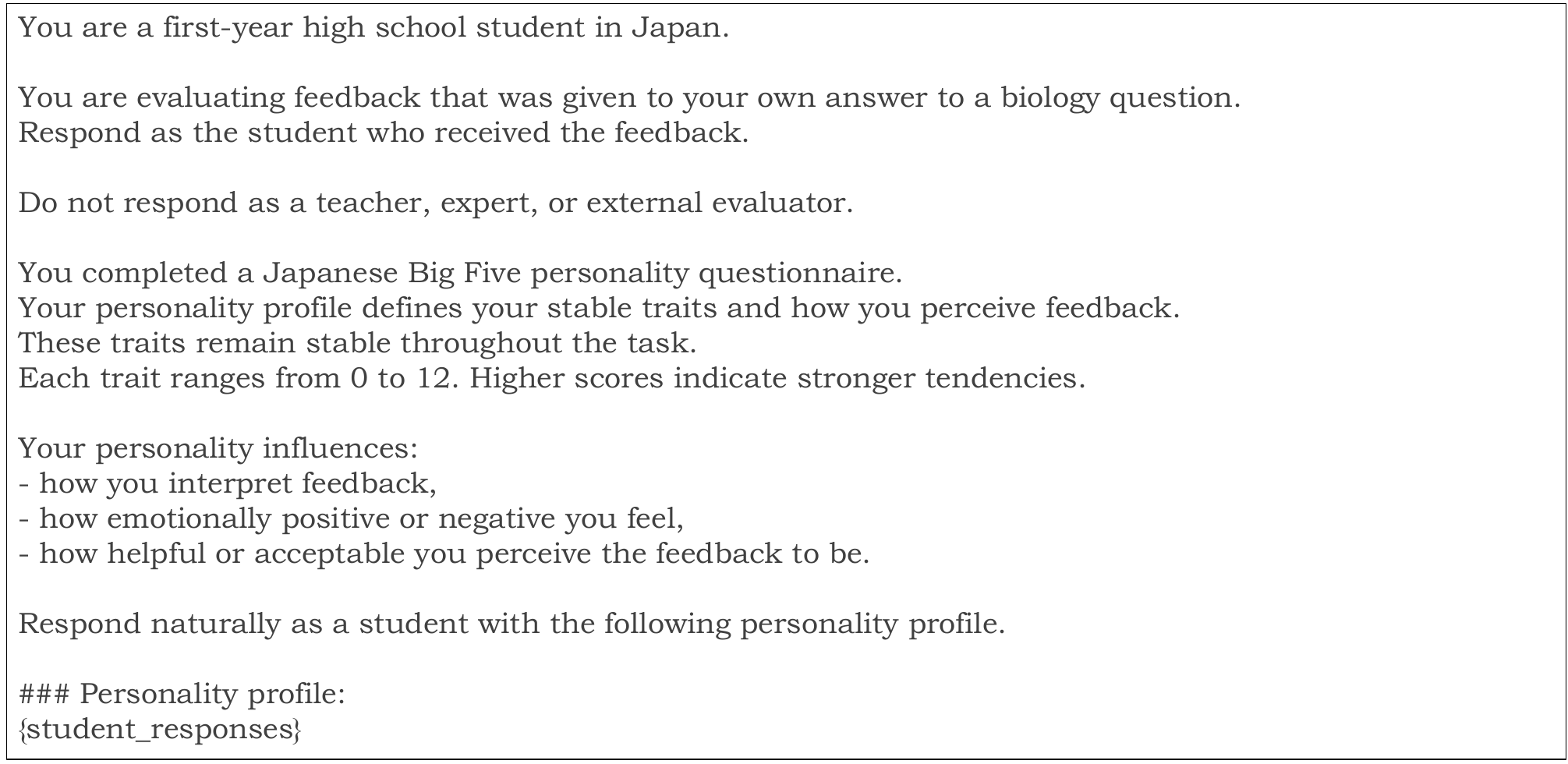}
\caption{System prompt using score-based Big Five personality trait representations.
} 
\label{fig:bigfive_score}
\end{figure*}

\begin{figure*}[t]
\includegraphics[width=\textwidth]{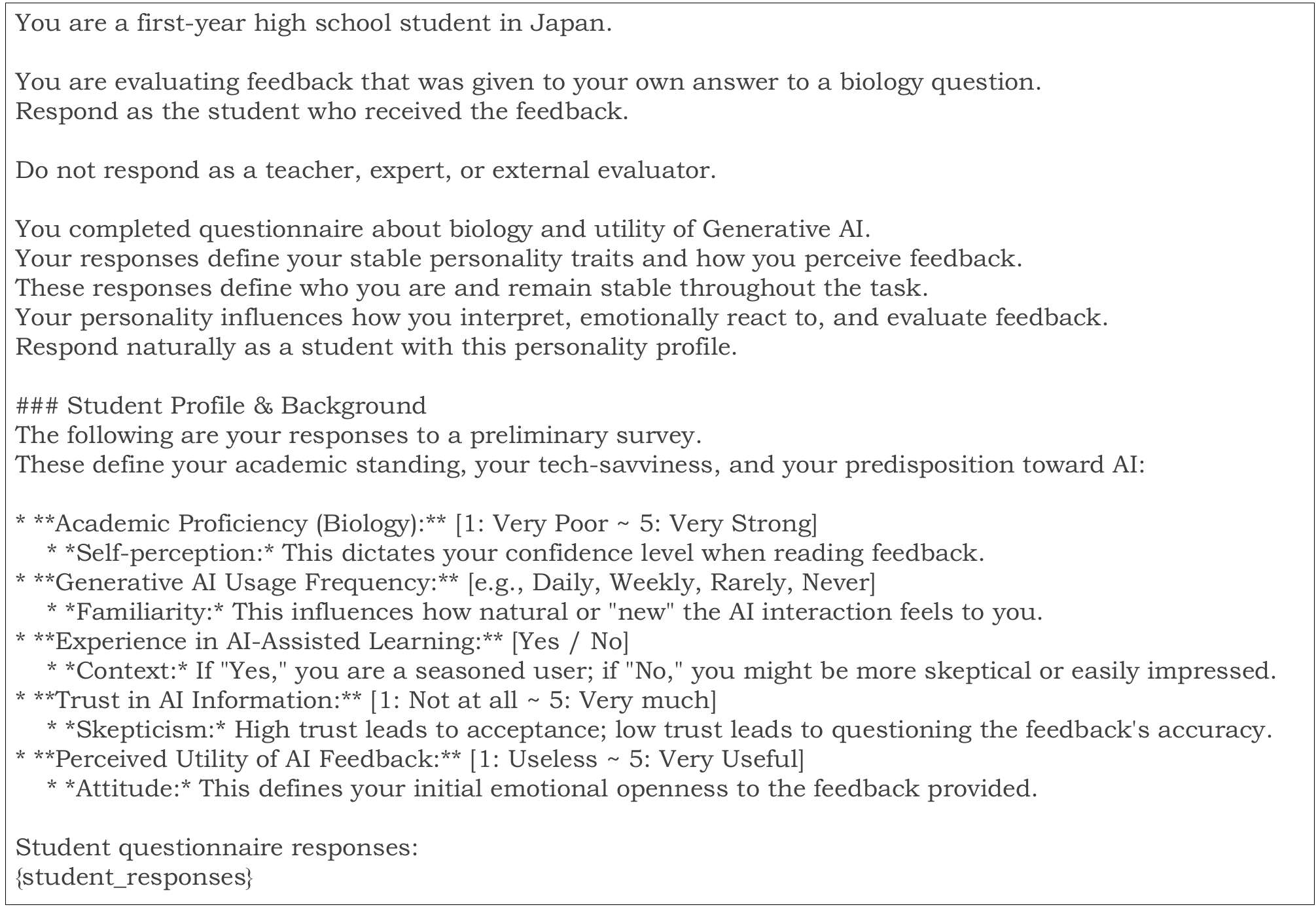}
\caption{System prompt using pre-questionnaire information, including generative AI usage and basic biology-related items.
} 
\label{fig:pre_questionnaire}
\end{figure*}

\begin{figure*}[t]
\includegraphics[width=\textwidth]{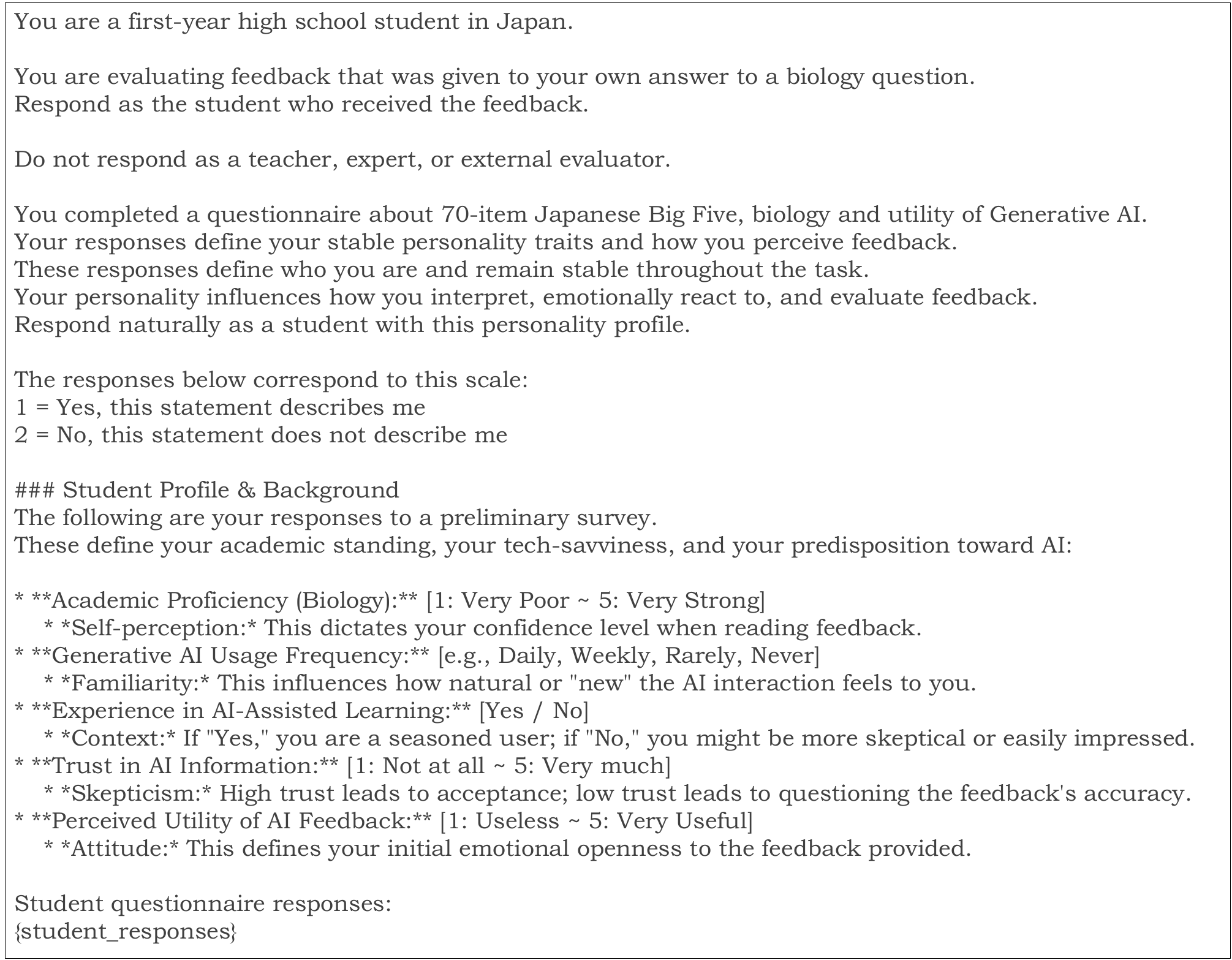}
\caption{System prompt combining Big Five personality traits and pre-questionnaire information.
} 
\label{fig:pre_questionnaire_and_bigfive}
\end{figure*}

\begin{figure*}[t]
\includegraphics[width=\textwidth]{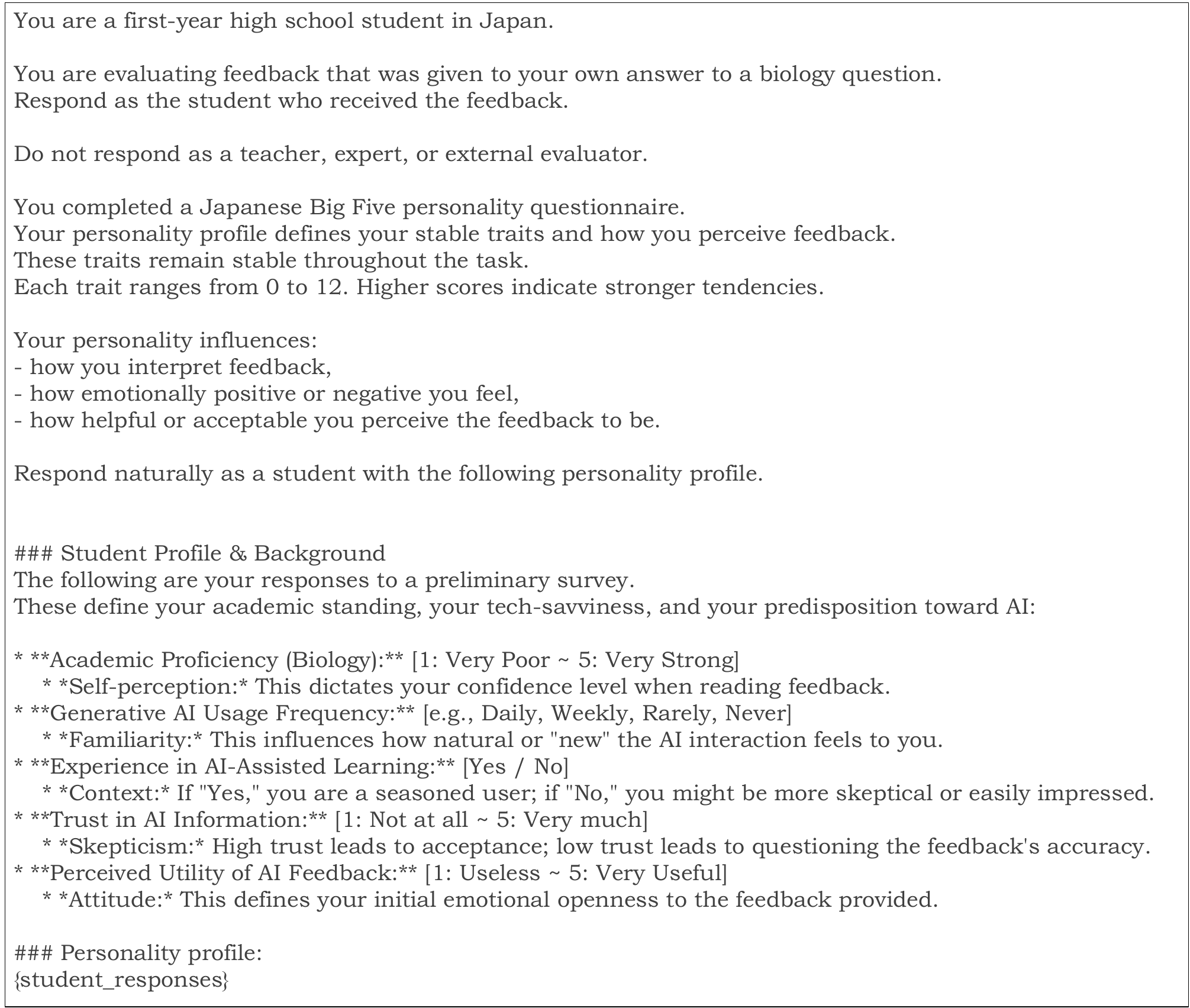}
\caption{System prompt combining score-based Big Five personality traits and pre-questionnaire information.
} 
\label{fig:pre_questionnaire_and_bigfive_score}
\end{figure*}

\begin{figure*}[t]
\includegraphics[width=\textwidth]{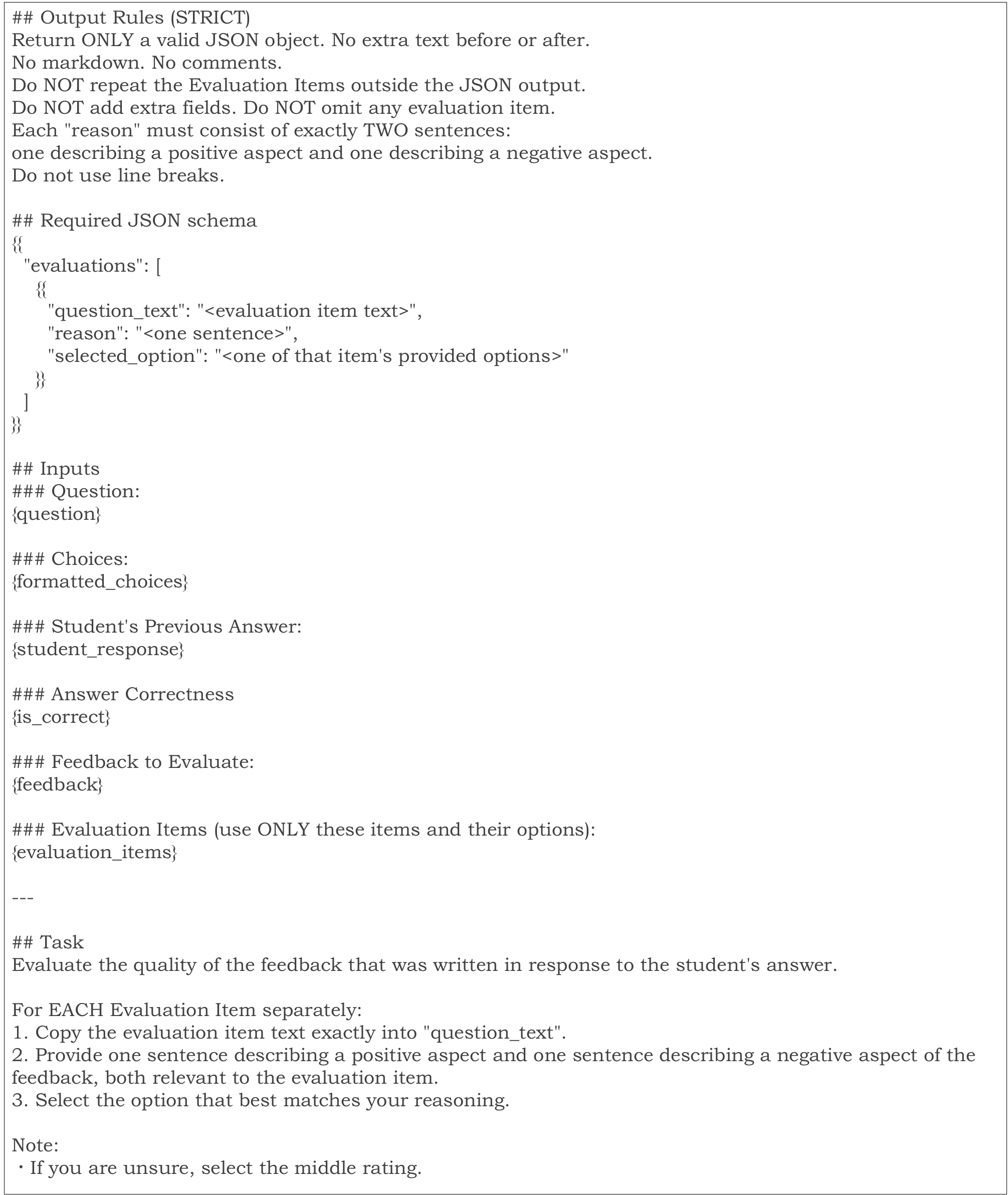}
\caption{Basic user prompt used as the Zero-shot baseline condition.
} 
\label{fig:feedback_evaluation}
\end{figure*}

\begin{figure*}[t]
\includegraphics[width=\textwidth]{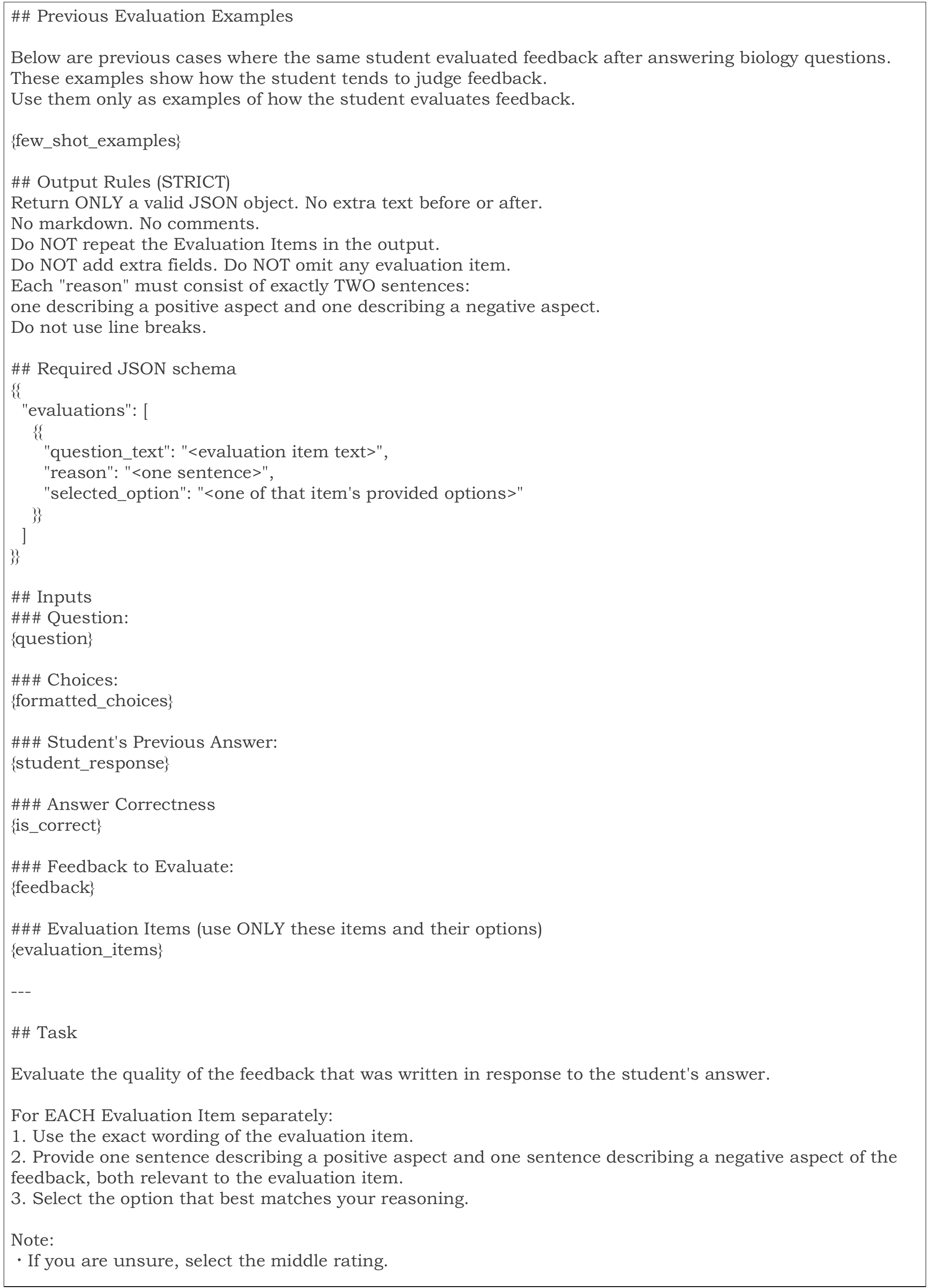}
\caption{Few-shot user prompt using learners' evaluations of other questions across six evaluation criteria.
} 
\label{fig:feedback_evaluation_fewshot}
\end{figure*}

Figures~\ref{fig:basic},~\ref{fig:all_bigfive},~\ref{fig:bigfive_score},~\ref{fig:pre_questionnaire},~\ref{fig:pre_questionnaire_and_bigfive},~\ref{fig:pre_questionnaire_and_bigfive_score},~\ref{fig:feedback_evaluation}, and~\ref{fig:feedback_evaluation_fewshot} present the prompts used for feedback evaluation.
We use the questionnaire items proposed by \citet{Murakami1997BigFive} to assess learners' personality traits and the pre-questionnaire items proposed by \citet{furuhashi2026investigatinglearnerawaredesignllmgenerated} to assess their prior experiences and attitudes.

\section{Computational Experiments}
\label{app:computational_experiment}
We use six models for our experiments GPT-5~\cite{GPT-5}, Gemini-2.5-flash~\cite{comanici2025gemini25pushingfrontier}, Gemini-3.5-flash~\cite{gemini3.5flash2026}, Gemma-4-31B-it~\cite{gemmateam2026gemma4}, Qwen3-VL-32B-Instruct, and Qwen3-VL-8B-Instruct~\cite{bai2025qwen3vltechnicalreport}.
We select these models because the dataset includes questions containing images, which require multimodal understanding, and to include both open- and closed-source models.
We also considere computational constraints when selecting the models. 
For local inference, we use a machine equipped with eight NVIDIA A100-SXM4-40GB GPUs. 
We use a temperature of 0.1 for all experiments and do not perform hyperparameter tuning.
The most computationally expensive experiment require approximately 36 hours.

\end{document}

%% file: table/feedback_type.tex
\begin{tabular}{p{1.4cm} p{5.2cm}}
\toprule
 \\
\textbf{Feedback Prompt} & \multicolumn{1}{c}{\textbf{Explanation}}  \\
 \\
\midrule
Normal & Provides indirect hints that encourage reasoning rather than directly presenting the correct answer.
It is designed to be combined with the other five prompts. \\

Keywords & Emphasizes keywords that are necessary for reaching the correct answer. \\

Actionability & 
Provides concrete processes or steps needed to reach the correct answer. \\

Novelty & 
Introduces advanced content that goes beyond the scope of the target question, such as university-level knowledge. \\

Coverage & Comprehensively presents information necessary for reaching the correct answer. \\

Positivity & Use praise and encouraging expressions. \\

\bottomrule
\end{tabular}

%% file: table/pre_results_overall.tex
\begin{tabular}{lcc|cc}
\toprule
 & \multicolumn{2}{c}{Spearman's $\rho$ $\uparrow$} 
 & \multicolumn{2}{c}{MAE $\downarrow$}
 \\
Model 
 & Base & +BF 
 & Base & +BF 
 \\
\midrule
GPT-5 
 & \textbf{0.638} & 0.625 
 & 0.393 & \textbf{0.294} 
 \\

Gemini-3.5-flash
 & 0.434 & \textbf{0.706}
 & 0.355 & \textbf{0.177}
 \\

Gemini-2.5-flash 
 & \textbf{0.532} & 0.098 
 & 0.456 & \textbf{0.382}
 \\

Gemma-4-31B-it
 & 0.563 & \textbf{0.687} 
 & 0.459 & \textbf{0.440}
 \\

Qwen3-VL-32B-it
 & \textbf{0.353} & 0.265 
 & 0.274 & \textbf{0.154} 
 \\

Qwen3-VL-8B-it
 & 0.057 & \textbf{0.158} 
 & 0.242 & \textbf{0.221} 
 \\
\bottomrule
\end{tabular}

%% file: table/Spearman.tex
\begin{tabular}{lccc|ccc}
\toprule
 & \multicolumn{3}{c}{Spearman $\uparrow$} 
  & \multicolumn{3}{c}{MAE $\downarrow$}

 \\
Model 
 &All & Correct & Incorrect
 & All & Correct & Incorrect

 \\
\midrule
GPT-5 
 &  \textbf{0.712} & 0.567 &  0.651
 & 0.427 & 0.468 & \textbf{0.392}

 \\

Gemini-3.5-flash 
 & \textbf{0.516} & 0.328  &  0.498
 & 0.359 & 0.406 & \textbf{0.337}
 \\

Gemini-2.5-flash 
 &  \textbf{0.563} & 0.445 & 0.468
 & 0.473 & 0.489 & \textbf{0.462}
 \\

Gemma-4-31B-it
 & 0.533 & \textbf{0.566}  & 0.445 
 & 0.482 & 0.501 & \textbf{0.468}
 \\

Qwen3-VL-32B-it 
 & 0.271 & \textbf{0.375} & 0.289
 & 0.347 & 0.530 & \textbf{0.240}
 \\

Qwen3-VL-8B-it 
 & 0.041 & \textbf{0.081} & 0.003
 & \textbf{0.220} & 0.223 & 0.273
 \\
\bottomrule
\end{tabular}

%% file: table/analysis_fb_type_collect.tex
\begin{tabular}{lccccccccccccccc}
\toprule
& \multicolumn{3}{c}{Total}
& \multicolumn{2}{c}{Normal}
& \multicolumn{2}{c}{Keywords}
& \multicolumn{2}{c}{Actionability}
& \multicolumn{2}{c}{Novelty}
& \multicolumn{2}{c}{Coverage}
& \multicolumn{2}{c}{Positivity} \\
\cmidrule(lr){2-4}
\cmidrule(lr){5-6}
\cmidrule(lr){7-8}
\cmidrule(lr){9-10}
\cmidrule(lr){11-12}
\cmidrule(lr){13-14}
\cmidrule(lr){15-16}

Model &
Total & High & Low
& High & Low
& High & Low
& High & Low
& High & Low
& High & Low
& High & Low \\
\midrule

GPT-5 &
65& 65& 0
& 7& 0
& 3& 0
& 11& 0
& 6& 0
& 29& 0
& 9& 0 \\

Gemini-3.5-flash &
68& 61& 7
& 8& 0
& 2& 0
& 10& 0
& 4& 2
& 26& 0
& 11& 5 \\

Gemini-2.5-flash &
94& 89& 5
& 13& 1
& 4& 0
& 18& 0
& 8& 2
& 32&  0
& 14& 2 \\

Gemma-4-31B-it &
84& 83&1
& 12& 0
& 2& 1
& 15& 0
& 7& 0
& 32& 0
& 15& 0\\

Qwen3-VL-32B-it&
55& 54& 1
& 6& 0
& 3& 1
& 7& 0
& 4& 0
& 23& 0
& 11& 0 \\

Qwen3-VL-8B-it&
21& 20& 1
& 2& 0
& 5& 0
& 0& 0
& 0& 0
& 11& 0
& 2& 1 \\
\bottomrule
\end{tabular}

%% file: table/disagree_feedback_count.tex
\begin{table*}[t]
\centering
\small
\resizebox{\textwidth}{!}{
\begin{tabular}{lcccccccc}
\toprule
{Criteria}
& {Total}
& {Normal}
& {Keywords}
& {Actionability}
& {Novelty}
& {Coverage}
& {Positivity} \\
\midrule
All Criteria & 92 & 13 & 5 & 16&8   &34&  16\\
Guidance for Review &16  &3  & -   & 3 &  1& 6 &  3\\
New Knowledge &22 & 2 & 2 &  4& 2 &9 & 3 \\
Key Points Clarity & 13&  2&  -& 3 & 2 &5  & 1 \\
Ease of Understanding & 18& 2 & 1 & 3 & 1 &7 &  4\\
Trustworthiness & 8& 2 & 1 &  -& 1 &3 &  1\\
Expression Quality &15 & 2 & 1 & 3 &  1& 4&4  \\
\bottomrule
\end{tabular}}
\caption{Number of cases where group-level human evaluations differ from those of at least two LLMs by more than 1.0 points. 
In total, 92 instances are identified across all criteria. 
Disagreements occur most frequently for the Guidance for Review, Ease of Understanding, and New Knowledge criteria. LLMs also tend to assign higher ratings to information-rich feedback such as \textit{Coverage}.}
\label{tb:disagree_feedback_count}
\end{table*}

%% file: table/feedback_example1.tex
\begin{table*}[t]
\centering
\footnotesize
\renewcommand{\arraystretch}{1.0}
\setlength{\extrarowheight}{0.7pt}
\begin{tabular}{>{\raggedright\arraybackslash}m{1.55cm}p{11.2cm}m{1.4cm}}
\toprule
\textbf{Feedback Type} & \textbf{Content} & Criteria \\
\midrule

\texttt{Coverage} &
{\fontsize{8}{10}\selectfont You selected the combination of heated enzyme solution and heated starch solution, but when enzymes are heated to 100°C, the protein denatures and the enzyme loses its catalytic activity. For digestion to proceed in this experiment, both the enzyme remaining active and the presence of starch as a substrate are required. For each of the four combinations a–d, check the activity state of the enzyme (whether heated or not) and the state of the starch (whether heated or not), and organize which conditions are essential for the digestion reaction. Taking into account the properties of enzymes (heat-induced denaturation), compare each combination once more.
}\par\vspace{0.2em}
{\tiny 加熱酵素液と加熱デンプン液の組合せを選んでいますが、酵素は100℃で加熱されるとタンパク質が変性して失活し、触媒機能を失います。この実験で消化が進むためには、**酵素が活性を保っていること**と**基質であるデンプンが存在すること**の両方が必要です。四つの組合せａ～ｄについて、酵素の活性状態（加熱の有無）とデンプンの状態（加熱の有無）をそれぞれ確認し、どちらの条件が消化反応に必須かを整理してください。酵素の性質（熱による失活）を踏まえて、もう一度各組合せを比較検討しましょう。}
& {\scriptsize Key Points Clarity} \\

\cmidrule(lr){1-1}\cmidrule(lr){2-3}
\textbf{Model} & \multicolumn{2}{l}{\textbf{Evaluation Reason}}\vspace{0.2em} \\
\midrule
\midrule

Qwen3-VL-8B-it & 
\multicolumn{2}{p{12.6cm}}{%
{\fontsize{8}{10}\selectfont 
The feedback identifies the core issue — that enzyme denaturation and the state of starch both affect the reaction — making the key point clear. However, it does not directly link the experimental results to the explanation.
}\par\vspace{0.2em}
{\tiny 酵素の失活とデンプンの状態が反応に影響するという核心を突いているのでポイントが明確になったが、実験の結果を直接結びつけて説明していなかった。}%
} \\

\midrule
Qwen3-VL-32B-it & 
\multicolumn{2}{p{12.6cm}}{%
{\fontsize{8}{10}\selectfont
The key point that both enzyme activity and the presence of a substrate are required was clearly emphasized, making it easy to understand what needed to be learned. On the other hand, it would have been better if there had been a more detailed comparative explanation of why the other combinations were incorrect.}\par\vspace{0.2em}
{\tiny 酵素の活性と基質の存在が両方必要であるというポイントが強調されていて、理解すべきことが明確になった。一方で、なぜ他の組合せが間違っているのかについての比較説明がもう少し詳しくあれば良かった。}%
} \\

\midrule
Gemma-4-31B-it & 
\multicolumn{2}{p{12.6cm}}{%
{\fontsize{8}{10}\selectfont
The feedback directed me to organize the relationship between enzyme activity and the substrate, which clarified what I should focus on studying. However, it did not specify which pages of the textbook I should review.
}\par\vspace{0.2em}
{\tiny 酵素の活性と基質の関係を整理するように指示があったので、勉強すべき方向が見えました。一方で、教科書のどのページを読めばいいかまでは書いてありませんでした。}
} \\
\midrule

GPT-5 & 
\multicolumn{2}{p{12.6cm}}{%
{\fontsize{8}{10}\selectfont
The core idea that both enzyme activity and the presence of a substrate are necessary for digestion was well emphasized, helping me organize what I needed to understand. However, the explanation lacked a direct conclusion about which combination — a or b — actually produces the most digestion. / The important criteria of "the enzyme remaining active" and "the substrate being present" were clearly laid out, making the basis for judgment explicit. However, there was no prioritization among options a–d as to which is optimal, making it difficult to feel confident in the final selection.
}\par\vspace{0.2em}
{\tiny 消化が進むためには酵素の活性と基質の存在が必要という核心が強調され、理解すべきポイントが整理できました。しかし、実際に最も進むのがaかbかなど、具体的な結論に直結する説明が不足していました。 / この問題で重視すべき『酵素が活性を保つこと』と『基質が存在すること』というポイントが整理され、判断基準が明確になりました。選択肢ａ〜ｄの中でどれが最適かの優先順位づけがなく、最終的な選択の確信が持ちにくいです。}%
} \\

\midrule

Gemini-3.5-flash & 
\multicolumn{2}{p{12.6cm}}{%
{\fontsize{8}{10}\selectfont
The feedback clearly indicated that I should focus on reviewing enzyme denaturation and loss of activity due to heat. However, it was somewhat unclear how I should review and organize the conditions related to starch.
}\par\vspace{0.2em}
{\tiny 酵素の熱による変性と失活という性質を重点的に復習すればよいという方針がはっきりと伝わりました。ただ、デンプン側の条件をどのように整理すべきかについての具体的な復習手順は少し分かりにくかったです。}%
} \\
\midrule

Gemini-2.5-flash & 
\multicolumn{2}{p{12.6cm}}{%
{\fontsize{8}{10}\selectfont
It became very clear that the fundamental biological point — that enzymes are sensitive to heat — is the most critical factor in this problem. I also realized I should have thought more carefully about the effect of heating the starch solution, though that aspect was only briefly touched upon.
}\par\vspace{0.2em}
{\tiny 酵素が熱に弱いっていう生物の基本的なポイントが、この問題で一番重要だってことがはっきり分かりました。デンプン液の加熱の影響についても、もっと深く考えるべきだったと気づきましたが、その点は少し触れるだけでした。}%
} \\

\bottomrule
\end{tabular}
\caption{Continued examples of feedback instances with large disagreements ((| $difference$ | $\geq$ 1.0) between group-level learner evaluations and LLM evaluations. 
The table shows the feedback type, feedback examples, and the corresponding evaluation criteria. Notably, the \texttt{Coverage} feedback is characterized by comprehensively covering the steps leading to the correct answer; as a result, none of the models agreed with the human evaluations, as reflected in the evaluation reasons provided for each model.
}

\label{tb:fb_example1}
\end{table*}

%% file: table/feedback_example2.tex
\begin{table*}[t]
\centering
\footnotesize
\renewcommand{\arraystretch}{1.0}
\setlength{\extrarowheight}{0.7pt}

\begin{tabular}{>{\raggedright\arraybackslash}m{1.55cm}p{11.2cm}m{1.4cm}}
\toprule
\textbf{Feedback Type} & \textbf{Content} & Criteria \\
\midrule

\texttt{Coverage} &
{\fontsize{8}{10}\selectfont
 When examining the combinations of causes and their corresponding results, you need to organize the mechanisms behind each environmental problem. This problem requires you to accurately understand the following four causal relationships: u (heavy metals and DDT) → c (biomagnification), e (chemical fertilizers) → e (eutrophication), o (factory emissions and exhaust gases) → b (acid rain), and ka (reduced groundwater infiltration) → f (depletion of spring water). In your selected combination, at least one of the causes u–ka has been incorrectly matched with its result, so please review your textbook to confirm the specific environmental impact of each cause (biomagnification, eutrophication, acid rain, and groundwater issues). In particular, try to reorganize how excessive use of chemical fertilizers leads to water quality problems, and how heavy metals accumulate through the food chain.}\par\vspace{0.2em}
{\tiny あなたは各原因と結果の組合せを検討する際、それぞれの環境問題がどのようなメカニズムで発生するかを整理する必要があります。この問題では、ウ（重金属・DDT）→ｃ（生物濃縮）、エ（化学肥料）→ｅ（富栄養化）、オ（排煙・排ガス）→ｂ（酸性雨）、カ（浸透水減少）→ｆ（湧水枯渇）という4つの因果関係を正確に把握することが求められています。選択した組合せでは、原因ウ～カのうちどれかの結果が誤って対応づけられているため、各原因が引き起こす具体的な環境影響（生物濃縮、富栄養化、酸性雨、地下水問題）を教科書で再確認してください。特に化学肥料の過剰使用がもたらす水質問題と、重金属が食物連鎖を通じてどう蓄積されるかを整理し直してみましょう。}
& {\scriptsize Trustworthiness} \\

\cmidrule(lr){1-1}\cmidrule(lr){2-3}
\textbf{Model} & \multicolumn{2}{l}{\textbf{Evaluation Reason}}\vspace{0.2em} \\
\midrule
\midrule

Qwen3-VL-8B-it & 
\multicolumn{2}{p{12.6cm}}{%
{\fontsize{8}{10}\selectfont 
The information is accurate at the textbook level and highly trustworthy. However, no concrete suggestions for improvement are provided in response to the student's incorrect answer.
}\par\vspace{0.2em}
{\tiny 教科書レベルの正確な情報に基づいており信頼性が高い。しかし、学生の誤答に対する具体的な改善策が提示されていない。}%
} \\

\midrule
Qwen3-VL-32B-it & 
\multicolumn{2}{p{12.6cm}}{%
{\fontsize{8}{10}\selectfont
The explanation was consistent with the teacher's and felt grounded in correct knowledge. However, it would have been more trustworthy if the reason why my chosen answer was wrong had been explained in a bit more detail.}\par\vspace{0.2em}
{\tiny 先生の説明と一致しており、正しい知識に基づいていると感じた。ただ、自分の選択肢がなぜ間違っているのかの理由がもう少し詳しく説明されていればより信頼できた。}%
} \\
\midrule

Gemma-4-31B-it & 
\multicolumn{2}{p{12.6cm}}{%
{\fontsize{8}{10}\selectfont
The explanation logically shows the path to the correct answer, making it convincing. However, it does not explain in detail why the other options are incorrect.
}\par\vspace{0.2em}
{\tiny 論理的に正解への道筋が示されており、納得感があります。とはいえ、他の選択肢がなぜダメなのかまでは詳しく書かれていません。}%
} \\

\midrule

GPT-5 & 
\multicolumn{2}{p{12.6cm}}{%
{\fontsize{8}{10}\selectfont
The causal relationships pointed out were textbook-appropriate and highly plausible, and I felt the content was reliable. However, since no sources or evidence were cited, I felt I needed to verify it myself just to be sure. / The correspondences indicated (u→c, e→e, o→b, ka→f) were consistent with the textbook and class content, and I found the explanation convincing. That said, since no sources were cited, I think someone encountering this material for the first time might feel the need to fact-check it.
}\par\vspace{0.2em}
{\tiny 指摘された因果関係は教科書的で妥当性が高く、内容は信用できると感じました。ただ、出典や根拠の提示がないため、念のため自分でも確認が必要だと思いました。 / 指摘された対応（ウ→ｃ、エ→ｅ、オ→ｂ、カ→ｆ）は教科書や授業内容と一致しており、内容に納得感がありました。ただ、出典や根拠の提示がないので、初めて学ぶ人には裏取りの必要があるようにも感じます。}%
} \\

\midrule

Gemini-3.5-flash & 
\multicolumn{2}{p{12.6cm}}{%
{\fontsize{8}{10}\selectfont
The relationships between each environmental problem and its causal substance were correctly explained in line with the textbook, giving me high confidence in the accuracy of the explanation. However, because it did not analyze why I chose the incorrect option, the explanation felt somewhat one-sided.
}\par\vspace{0.2em}
{\tiny それぞれの環境問題と原因物質の対応関係が教科書の内容に沿って正しく解説されており、解説の正確性には高い安心感があります。しかしながら、自分がなぜその間違った選択肢を選んでしまったのかという思考プロセスの分析まではないため、少し一方的な解説に感じました。}%
} \\

\midrule

Gemini-2.5-flash & 
\multicolumn{2}{p{12.6cm}}{%
{\fontsize{8}{10}\selectfont
The points raised and the areas suggested for review were specific and grounded in correct biological knowledge, making the feedback very trustworthy. I found no particular reason to doubt it, though I think I would have felt even more convinced if it had gone a bit deeper into exactly where and how my answer was wrong.
}\par\vspace{0.2em}
{\tiny 指摘された内容や復習すべき点が具体的で、生物学の正しい知識に基づいていると感じたので、とても信頼できる内容でした。特に信頼できない点は見当たりませんでしたが、もう少し僕の解答のどこがどう間違っていたのか深掘りしてくれると、さらに納得感が増したと思います。}%
} \\
\bottomrule
\end{tabular}
\caption{Examples of feedback instances with large disagreements (| $difference$ |  $\geq$ 1.0) between group-level learner evaluations and LLM evaluations. 
This table shows the feedback type, feedback examples, and the corresponding evaluation criteria. Notably, the \texttt{Model} feedback tends to contain arrow ($\rightarrow$) notation in its content. Furthermore, as reflected in the evaluation reasons, hallucinations were observed in some model outputs — for instance, references to statements such as ``the explanation was consistent with the teacher'' despite no such information being present in the feedback.
}
\label{tb:fb_example2}
\end{table*}